# Title: Computer-aided diagnosis of lung carcinoma using deep learning – a pilot study

Author: ZHANG LI[1], ZHEYU HU[2*], JIAOLONG XU[3*], TAO TAN[4**], HUI CHEN[5], ZHI DUAN[5], PING LIU[5], JUN TANG[6], GUOPING CAI[7], QUCHANG OUYANG[2], YULING TANG[5**], GEERT LITJENS[8], QIANG LI[9**]

**Affiliation:**

1. College of Aerospace Science and Engineering, National University of Defense Technology, Changsha 410073, China

2. Department of Breast Medical Oncology and Central Laboratory, Affiliated Cancer Hospital of Xiangya School of Medicine, Central South University, Changsha 410000, China

3. Computer Vision Center, Autonomous University of Barcelona, Spain

4. Department of Biomedical Engineering, Eindhoven University of Technology, Eindhoven 5600 MB, The Netherlands

5. The First Hospital of Changsha City, Changsha 410000, China

6. The First People's Hospital of Foshan, Foshan 528000, China

7. Department of Pathology, Yale School of Medicine, New Haven, CT06519, US

8. Department of Pathology, Radboud University Medical Center, Nijmegen, The Netherlands

9. Department of Respiratory Medicine, ShangHai Ease Hospital, Tongji University, 310000, China.

*ZHEYU HU, JIAOLONG XU and ZHANG LI have equal contribution to this work

**Corresponding author: TAO TAN (email:tao.tan911@gmail.com); YULING TANG(email:tyl71523@sina.com)

**Abstract**

**Aim**: Early detection and correct diagnosis of lung cancer are the most important steps in improving patient outcome. This study aims to assess which deep learning models perform best in lung cancer diagnosis.

**Methods**: Non-small cell lung carcinoma and small cell lung carcinoma biopsy specimens were consecutively obtained and stained. The specimen slides were diagnosed by two experienced pathologists (over 20 years). Several deep learning models were trained to discriminate cancer and non-cancer biopsies.

**Result**: Deep learning models give reasonable AUC from 0.8810 to 0.9119.

**Conclusion**: The deep learning analysis could help to speed up the detection process for the whole-slide image (WSI) and keep the comparable detection rate with human observer.

**Introduction**

Lung cancer is the top cause of cancer-related death in the world. According to 2009-2013 SEER (Surveillance, Epidemiology and End Results) database, the 5-year survival rate of lung cancer patients is only about 18% (https://seer.cancer.gov/statfacts/html/lungb.html) [1]. For early stage, resectable cancer, the 5-year survival rate is about 34%, but for unresectable cancer, the 5-year survival rate is less than 10%. Therefore, early detection and diagnosis of lung cancer is one of the most important steps in improving patient outcome. According to NCCN (National Comprehensive Cancer Network) guidelines, for image-suspected tumors, histopathological assessment of biopsies obtained via fiberoptic bronchoscopy should be performed for early diagnosis [2-3].

Assessment of biopsy tissue by a pathologist is the golden standard for lung cancer diagnosis, however, the diagnostic accuracy was less than 80% [4]. The major histological subtypes of malignant lung disease are squamous carcinoma, adenocarcinoma, small cell carcinoma and undifferentiated carcinoma. Correctly assessing these subtypes on biopsy is paramount for correct treatment decision. However, the number of qualified pathologists is too small to meet the huge clinical demands, especially in countries such as China with a big population of lung cancer patients. Automatic assessment of lung biopsies by an artificial intelligence (AI) system might be able solve this problem efficiently.

In recent years, AI techniques flourished rapidly in the field of medical diagnosis. In 2016, convolutional neural networks were used to successfully detect melanoma lesions with 92% accuracy [5]. In breast cancer, tetrapolar impedance measurements (TPIM), with feature extraction by machine learning, reached an effective diagnosis rate of 84% [6]. In classifying breast cancer cells, an AI technique reached 95.34% [7]. Currently, the application of AI in lung cancer diagnosis has focused mostly on radiology, for example using radiomics [8-9]. Radiomics is a process that automatically quantifies radio-phenotypic characteristics by agglomerating imaging-derived features [8]. With engineered CT image analytics, radiomics signatures could distinguish EGRF- and EGFR+ lung cancers, and EGFR+ and KRAS+ lung cancers [8]. Convolutional neural networks (CNN) were capable of reached an accuracy rate of 86.4% in classifying pulmonary nodules [9]. However, application of these AI techniques has not yet found its way to histopathological analysis of lung cancer.

However, in past decades, researchers have developed multiple automatic or semi-automatic quantification models to objectively evaluate pathological figures in other types of cancer [5, 10]. Traditional research steps include digitization of histopathological specimens into whole-slide images (WSIs), lesion segmentation, feature extraction, and classifier training. In traditional AI techniques, the extracted features were usually hand-crafted, such as shape, border, color change and texture descriptors [11-13]. Most recently, deep learning (DL) algorithms, especially convolutional neural networks (CNN), have been successfully applied to digital pathology image analysis [14-15]. Compared to traditional machine learning algorithms, deep learning methods do not require any handcrafted features. Training deep learning models often requires a lot of annotated images, which can be difficult to acquire in the medical field. However, a lot of CNNs were designed for natural image analysis, which, via transfer learning, could help researchers solve problems in medical images.

CNN-based algorithms have been used from the cell level to the WSI level. For the cell level, several DL algorithms that are based on CNNs were used in mitosis detection [16-18] and nucleus detection [19-20]. Classifying nuclei was also performed with CNNs, in which data augmentation techniques were used to increase the accuracy [21-22]. Larger objects, such as glands, are also important for pathologists to assess grade of certain cancers. Deep learning algorithms have also been applied to segment such objects from WSIs. Several authors have tried different approaches based on contour information [23], handcrafted features [24] and multi-loss [25], which were incorporated in conventional CNNs to obtain reliable segmentation results.

At the WSI level, a CNN with only 3-layers was first introduced to detect invasive ductal breast carcinoma and showed a comparable result (65.40% accuracy) with hand-crafted features based classifiers [26]. More complex CNN were also used in the detection of prostate and breast cancer [27]. Instead of only using the original RGB image, other image representation such as the magnitude and phase of shearlet coefficient images can be fed into a CNN [28]. This method surpassed the handcrafted features based detection methods and gave higher detection rates (86% accuracy) of breast cancer than only using RGB image (71% accuracy). CNNs were also used to extract features for better colon cancer classification and colon cancer prediction [29]. More deeper CNN, such as GoogLeNet [30], AlexNet [31], VGG and ResNet [32], were used for breast cancer classification [33] and prostate cancer prediction [34].

Several Grand Challenges in Medical Imaging also greatly advanced the pathology image analysis community, such as mitosis detection challenges in ICPR 2012[1] CAMELYON16[2] and CAMELYON17[3] for identifying breast cancer metastases. In particular, the CAMELYON16 was the first challenge offering the WSIs with large amount of annotations, which is essential for training larger CNNs such as ResNet.

---

[1] http://people.idsia.ch/~juergen/deeplearningwinsMICCAIgrandchallenge.html

[2] https://camelyon16.grand-challenge.org/

[3] https://camelyon17.grand-challenge.org/

However, even after assessing recent review papers [14-15], we found no papers discussing the applications of CNNs to histopathological images of lung cancer. Furthermore, no public datasets of WSI were available to evaluate such algorithms.

To our best knowledge, we are the first group address this issue. Therefore, in this study, we collected pathological WSIs from 40 lung cancer patients. WSIs from 7 of them were excluded because their lung samples were brushing cells obtained by using bronchoscopy. These samples were basically cytological specimen, and show a quite different appearance compare to other 33 samples that were taken by surgery. Experienced pathologists (over 20 years of experience) identified the cancer regions on each slide. We then compared several CNN-based algorithms in performing lung cancer diagnosis.

**Methods**

*Patient recruitment*

33 lung patients were recruited in this study. All patients were treated at the Department of Pulmonary Oncology in the First Hospital of Changsha, from January 2016 to November 2017. According to American Joint Committee on Cancer (AJCC) staging system, patients firstly diagnosed with lung/bronchus cancer (site: C34.1-C34.9; histology type: adenocarcinoma, squamous cell carcinoma and small cell carcinoma) were recruited. Other inclusion criteria included: 1) pathologically confirmed patients with surgery biopsy maintained; 2) no radiotherapy before surgery; 3) aged between 30 and 90 yr; 4) detailed clinical information is available. The exclusion criteria were: 1) multiple primary cancers; 2) metastatic lung cancer; 3) patients with immune-deficiency or organ-transplantation history; 4) patients without detailed clinical information; 5) patients who did not provide informed consent. This study was approved by the Ethics Committee of the First Hospital of Changsha. Informed consent was obtained from each patient before study. Basic demographic and clinical information for each patient, such as age, pathology, laterality, stage, imaging records, treatment history were collected.

*Image acquisition and pre-processing*

Histological slides (3 slides per patient) were stained with hematoxylin & eosin (H&E). The stained tissue slides were then scanned by an automated microscope (Olympus VS120) with at objective magnifications of 20x. One experienced pathologist annotated the cancer regions, see **Fig.1**. Similar to the [33], the 20x images were cropped into small patches with size of 256*256 pixels. The cropped patches are shown in **Fig.2**. One can see that the patch colors were quite different even among the patches from normal tissue due to the staining variability. The appearance of the cancer regions is also quite different because of the different cancer types. For instance, **Fig 2.(A) and (B)** show small-cell lung cancer and **Fig 3.(C) and (D)** show non-small cell lung cancer tissues.

*Experiments setup*

We randomly split the data into training set and test set with 26 and 7 slides, respectively. For each slide, image patches of 256x256 pixels are cropped with stride of 196 pixels, to ensure sufficient overlapping between adjacent patches. Finally, there are around 80000 small image patches in training set and 30000 small image patches in test set. Since the annotations were only assigned cancer or not cancer at this moment, our problem is defined as binary classification. The classification is given at the patch level in the end.

*Traditional algorithm with GLCM and SVM*

Texture analysis with GLCM (Gray-level Co-occurrence Matrix) has been widely used in the analysis of cancer pathology [35-36]. Cancer patches and normal tissue patches were normalized by using median and quartiles as following equation:

$$Norm(I) = \frac{2*(I - quantile(I, 0.5))}{\|quantile(I, 0.75) - quantile(I, 0.25)\|}$$

where the *quantile(I,x)* calculates the *x* quauntile of patch *I*. Then normalized figures were divided into small segments of size of 7×7 pixels. For each figure, texture characters were extracted by using the mean, variance, homogeneity, contrast, dissimilarity, entropy, second moment and correlation of the small segments. A support vector machine (SVM) classifier was then trained using mean and variance features to discriminate benign and malignant tumors.

*Convolutional neural networks*

Convolutional neural networks (CNN) are a type of the neural networks that is particularly suited for image analysis. It has been, for example, successfully used for image classifications [31, 37-38]. A typical CNN architecture contains convolutional, pooling and fully-connected layers. Relatively novel techniques such as batch normalization [39], dropout [40] and shortcut connections [32] can additionally be used to increase classification accuracy.

*Patch-based classification by CNN*

MXNet [41] package is used for training our deep learning models. We have tested several popular CNN architectures for the patch-based classification: AlexNet [31], VGG[42], ResNet [32] and SqueezeNet [43]. All the networks were pre-trained on ImageNet [44], the current largest image classification dataset in computer vision. AlexNet was the winner of ImageNet competition 2012 [44]. VGG [42], was an even deeper CNN, and won the ImageNet competition 2014. Based on the concepts of shortcut connections and residual representations, ResNet allowed exploring even deeper architectures(from 18-layer to 152-layers) and won ImageNet competition 2015. Compared to the aforementioned three networks, SqueezeNet is a smaller network with less parameters. The small network is easier to deploy on the hardware with limited memory.

We compared two types training schemes: training from the scratch and fine tuning pre-trained networks. The pre-trained weights on ImageNet are used as an initialization in our finetuning experiments. We fixed the learning rate=0.00001, weighted decay rate=0.0001, epoch=10 and batch size = 64 for all four methods. VGG-16 and ResNet-50 were selected to represent VGG and ResNet architectures for the experiments. The Adam algorithm [45] was used to optimize the weights.

**Results and Discussion**

*Patch based classification*

The merits of the algorithms were assessed for discriminating cancer patches and normal patches. Receiver operating characteristic (ROC) analysis was performed at patch level and the measure used for comparing SVM based method and CNNs was the area under the ROC curve (AUC). The results of SqueezeNet, ResNet-50, Alexnet and VGG-16 are shown in **Fig.3** and **Fig.4. Fig 3.** Shows the result for training from scratch and Fig 4 for fine tuning the pre-trained networks. One can see from Fig.3 that the traditional learning method, namely GLCM+SVM, gave the lowest AUC. One can also see that training from scratch showed higher AUC than fine tuning the whole network except for ResNet-50. One can also see that AlexNet gave the highest AUC in training from scratch strategy. The true positive rate with respect to 3 typical false positive rates, at 0.05, 0.5 and 0.95, are also shown in Table I and Table II.

*Slide based cancer region detection and comparisons with human annotations*

We conducted the slide-based cancer region detection by combining all patch based classification result from the previous section. The heatmaps for each WSI were calculated for this task (see **Fig.5**).

Within the annotated area, one can see that most of the region is with high cancer probability. One can also see some "false" negative region within the annotated area too (e.g. the place pointed out by the green arrow). However, when we zoomed into those regions (see **Fig. 5(c) and (e)**), we could see that the "false" negative regions are actually the true negative regions. This observation was confirmed by the same pathologist and can be found in several slides. It means that those regions were predicted correctly by the DL based model. During the training stage, these "false" positive annotated regions could be seen as the label noise for the model. Therefore, the DL model is robust to label noise. The DL model could also separate the boundary between the cancer region and normal region (see **Fig. 5(d) and (f)**). We may need to introduce another DL model to reduce the false positive as one did in [33].

**Conclusion**

In this paper, we have tested several DL models for lung cancer diagnose using WSI in histopathology. The preliminary results show that DL methods have potential for lung cancer diagnosis. However, the comparing our results to those in other reported cancer diagnosis systems based on deep learning, we have lower AUCs, showing the challenges of the lung cancer diagnose [33]. This may be due to the large variation of the patterns between different slides which result in discrepancy between training and testing data, which is an inherent limitation of a small dataset size. Unlike fine-tuning for other computer vision tasks, our models have not seemed benefit too much from the imageNet pre-trained models. This could be because our domain is inherently different from the ImageNet domain, and the weights learned from imageNet actually have little contribution to our final model However, how to select the best architecture is still an open question. The non-precise annotation from the pathologist also degraded the detection accuracy, as training deep learning methods with noisy labels results in sub-optimal learning.

In future work, on one hand, we will collect more training data to cover the large variation of data distribution, on the other hand we will investigate other techniques, e.g., domain adaptation [31] or sequential tuning [46] to address the distribution discrepancy and increase the AUCs. We will also create more detailed annotations and focus on separating the different types of lung cancer. The dataset used in the paper will be publicly available upon acceptance of the paper and put in a place-holder link.

**Funding**

This work partially funded by Health and Family Planning Commission of Hunan Province Foundation (B20180393 and C20180386) and Changsha City Technology Program (kq1701008)

**Acknowledgments**

The authors would like to thank Tao Xu, Jun XU, Shanshan Wan, Ke Lou, Hui Li, Keyu Li and Yusheng Yan for collecting all the images.

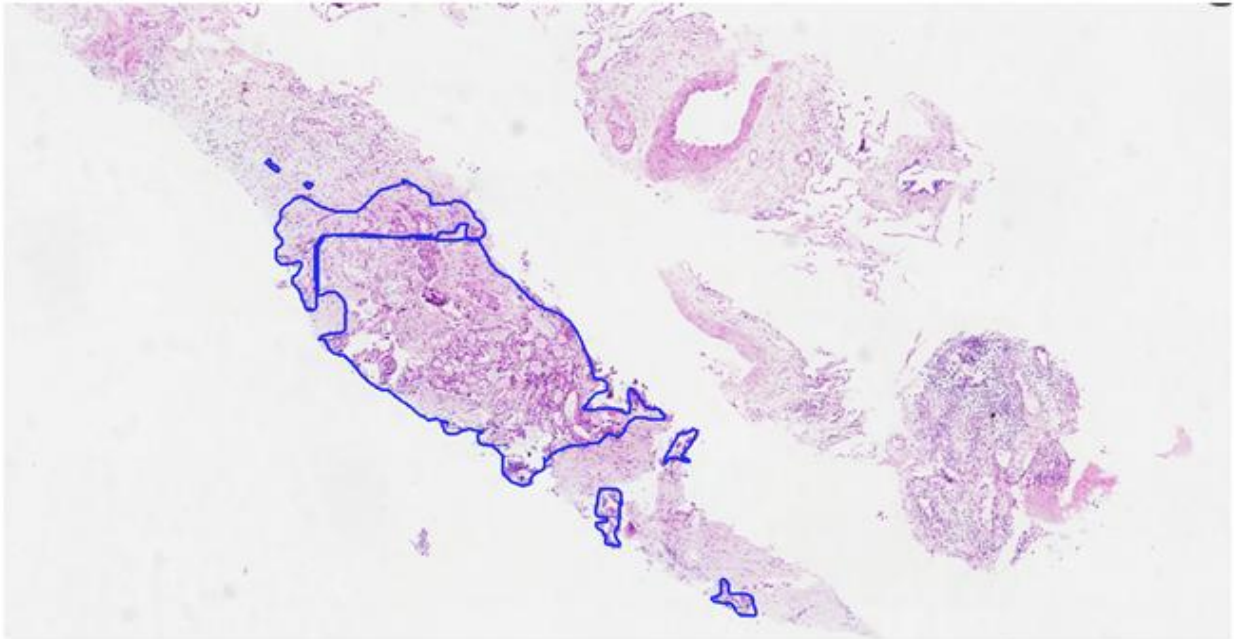

**Figure 1.** Pathological WSI with annotations for cancer regions (the image is zoomed in 3x for better visualization).

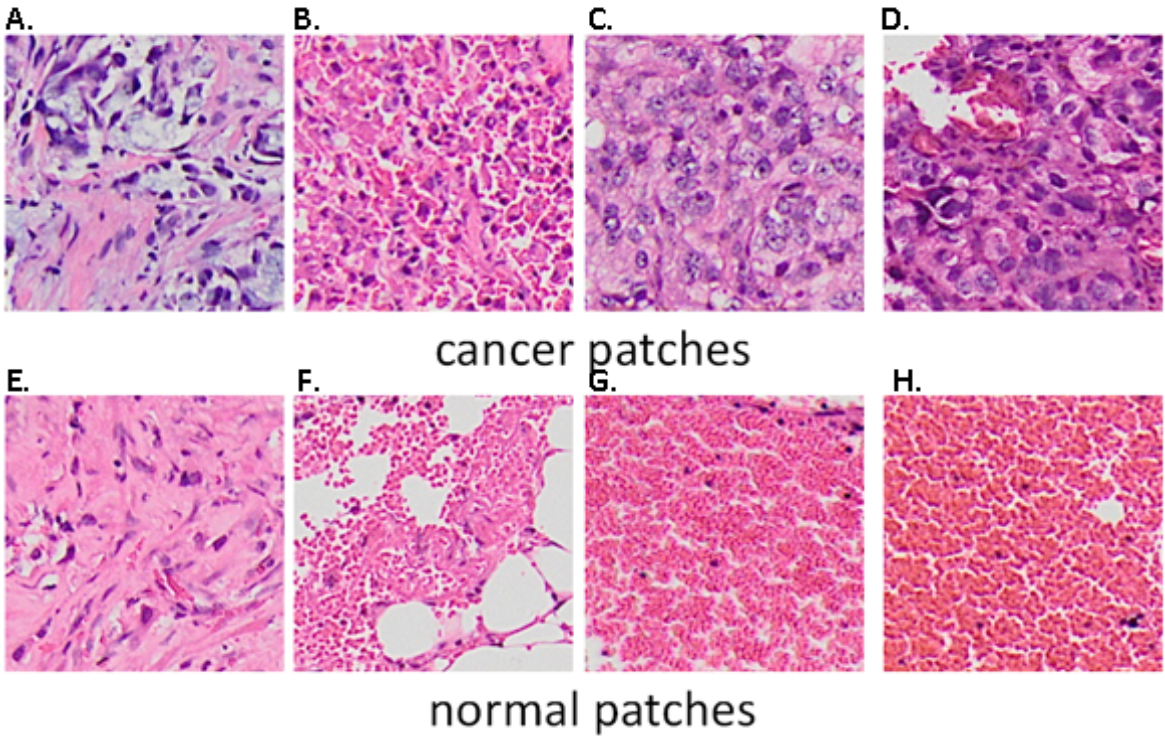

**Figure 2.** examples of tumor patches and normal patches.

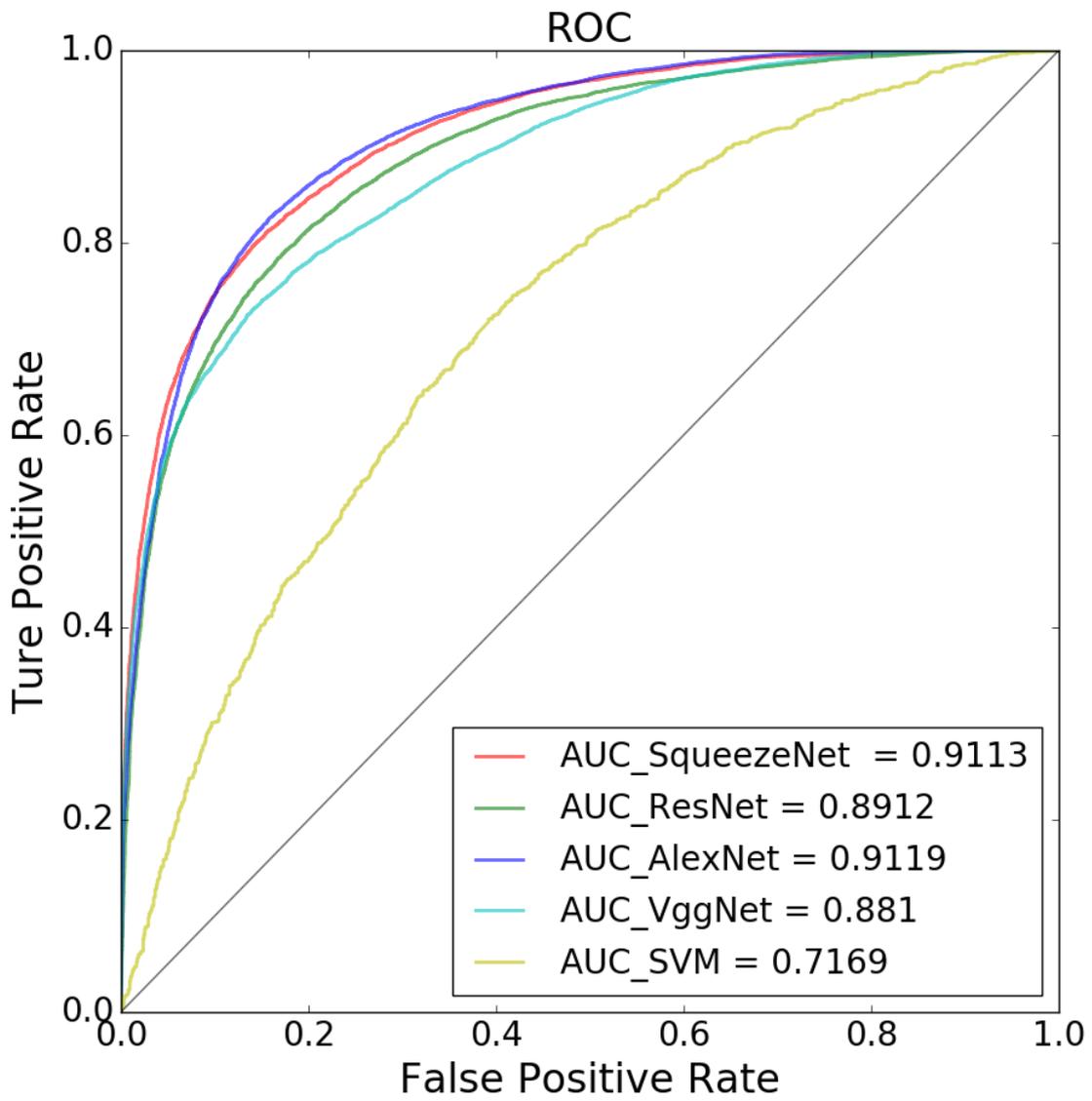

**Figure 3.** ROC of training from scratch

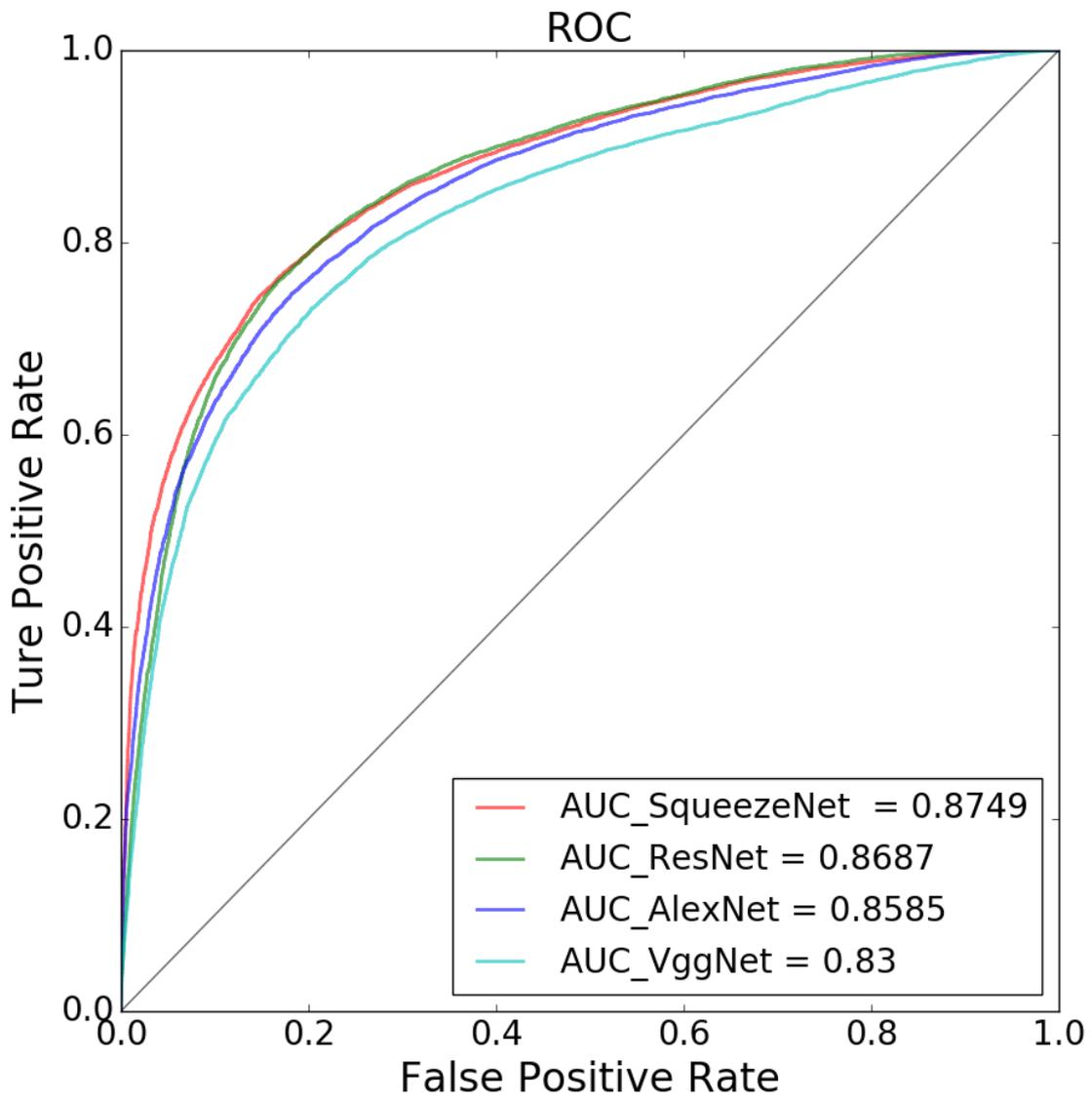

**Figure 4.** ROC of fine tuning for whole networks.

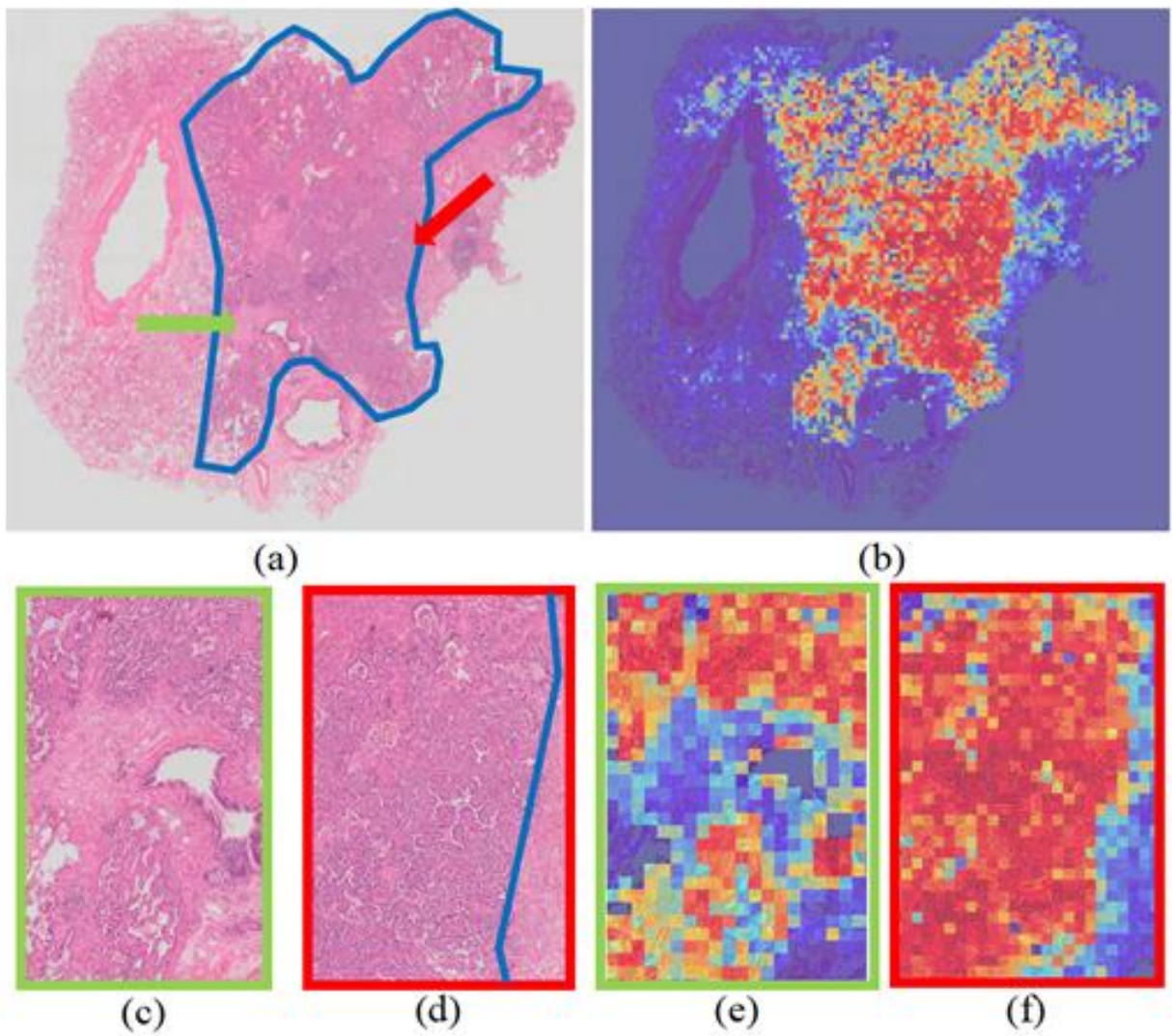

**Figure 5.** Visualization of cancer region detection. (a) original image with annotations from one experienced pathologist (blue polygon); (b) the heatmap over the original image; (c) and (d) the representative sub-regions from the original image that pointed out by green and red arrows; (e) and (f) the corresponding heatmaps of sub-regions (c) and (d). The higher cancer probability of the patch, the hotter (more reddish) color for that patch was.

TABLE 1 The true positive rate of training from scratch with respect to different false positive rate.

|  | FP@0.05 | FP@0.1 | FP@0.5 |
|---|---|---|---|
| SqueezeNet | 0.6344 | 0.7460 | 0.9688 |
| ResNet | 0.5779 | 0.6948 | 0.9542 |
| AlexNet | 0.6039 | 0.7455 | **0.9704** |
| VggNet | 0.5840 | 0.6760 | 0.9423 |
| SVM | 0.1643 | 0.3011 | 0.8069 |

TABLE 2 The true positive rate of fine tuning with respect to different false positive rate.

|  | FP@0.05 | FP@0.1 | FP@0.5 |
|---|---|---|---|
| SqueezeNet | 0.5644 | 0.6735 | 0.9268 |
| ResNet | 0.4863 | 0.6582 | **0.9307** |
| AlexNet | 0.5060 | 0.6330 | 0.9176 |
| VggNet | 0.4431 | 0.5923 | 0.8899 |